\documentclass[runningheads]{llncs}
\usepackage{graphicx}

\usepackage{tikz}
\usepackage{comment}
\usepackage{amsmath,amssymb} 
\usepackage{color}

\usepackage[accsupp]{axessibility}  

\usepackage{graphicx}
\usepackage{amsmath}
\usepackage{amssymb}
\usepackage{booktabs}
\usepackage{mathtools}
\usepackage{bm}
\usepackage[pagebackref,breaklinks,colorlinks]{hyperref}
\usepackage[capitalize]{cleveref}

\usepackage{caption}

\begin{document}
\pagestyle{headings}
\mainmatter
\def\ECCVSubNumber{7151}  

\title{Should All Proposals be  Treated Equally in Object Detection?} 

\titlerunning{ECCV-22 submission ID \ECCVSubNumber} 
\authorrunning{ECCV-22 submission ID \ECCVSubNumber} 
\author{Anonymous ECCV submission}
\institute{Paper ID \ECCVSubNumber}

\titlerunning{DPP}
%
\author{Yunsheng Li\inst{1,2} \and Yinpeng Chen\inst{1} \and Xiyang Dai\inst{1} \and Dongdong Chen\inst{1} \and Mengchen Liu\inst{1} \and Pei Yu\inst{1} \and Ying Jin\inst{1} \and Lu Yuan\inst{1} \and Zicheng Liu\inst{1} \and Nuno Vasconcelos\inst{2}}
\authorrunning{Y. Li et al.}
%
\institute{Microsoft Corporation, Redmond WA 98052, USA \and
UC San Diego, La Jolla, CA 92093, USA\\
\email{\{yunshengli,yiche,xidai,dochen,mengcliu,pei.yu,\\ying.jin,luyuan,zliu\}@microsoft.com, nvasconcelos@ucsd.edu }}
\maketitle

\begin{abstract}
The complexity-precision trade-off of an object detector is a critical problem for resource constrained vision tasks. Previous works have emphasized detectors implemented with efficient backbones. The impact on this trade-off of proposal processing by the detection head is investigated in this work. It is hypothesized that improved detection efficiency requires a paradigm shift, towards the unequal processing of proposals, assigning more computation to good proposals than poor ones. This results in better utilization of available computational budget, enabling higher accuracy for the same FLOPS. We formulate this as a learning problem where the goal is to assign operators to proposals, in the detection head, so that the total computational cost is constrained and the precision is maximized. The key finding is that such matching can be learned as a function that maps each proposal embedding into a one-hot code over operators. While this function induces a complex dynamic network routing mechanism, it can be implemented by a simple MLP and learned end-to-end with off-the-shelf object detectors. This {\it dynamic proposal processing} (DPP) is shown to outperform state-of-the-art end-to-end object detectors (DETR, Sparse R-CNN) by a clear margin for a given computational complexity. Source code is at \href{https://github.com/liyunsheng13/dpp}{https://github.com/liyunsheng13/dpp}
\end{abstract}

\section{Introduction}
\label{sec:intro}
Object detection is a challenging but fundamental task in computer vision, which aims to predict a bounding box and category label for each object instance in an image. A popular strategy, introduced by the Faster RCNN \cite{ren2015faster}, is to rely on a backbone network to produce a relatively large set of object proposals and a detection head to derive a final prediction from these. 
Since then, the design trend for this two-stage detection framework, e.g. the path Faster RCNN \cite{ren2015faster} $\rightarrow$ Cascade RCNN \cite{cai2018cascade} $\rightarrow$ DETR \cite{carion2020end} $\rightarrow$ Sparse RCNN \cite{sun2021sparse}, has been to sparsify the proposal density. Recent approaches, such as the Sparse R-CNN~\cite{sun2021sparse}, successfully reduce the thousands of proposals of the Faster RCNN \cite{ren2015faster} to a few hundred. 
However, because the per proposal computation of the detection head is substantially increased by the use of a much more complicated architecture, the overall computational benefits of reducing the number of proposals are limited. While the aggregate effect has been to make detectors more efficient, in general, these approaches are still not suitable for use with lighter backbones, since the head complexity becomes a larger fraction of the overall computation.

\begin{figure}[t]
    \centering
    \includegraphics[width=0.98\linewidth]{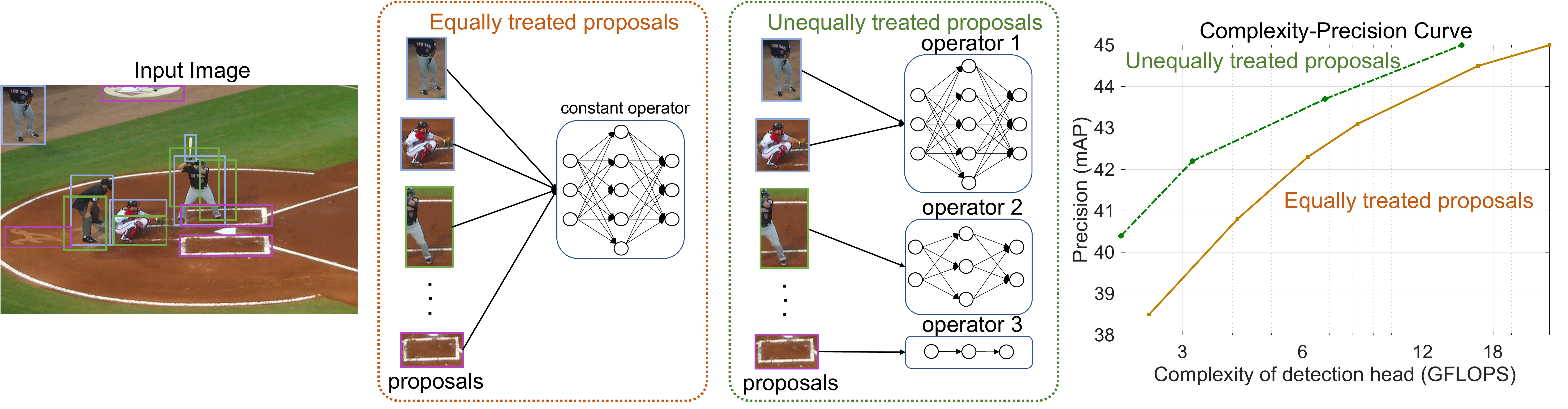}
    \caption{Existing object detectors treat proposals equally, applying the same operator to all proposals. Dynamic Proposal Processing (DPP) instead argues for an unequal treatment, by learning to dynamically assign different proposals to operators of different complexities. This enables the allocation of more (less) computation to high (low) IoU proposals and enables improved complexity-precision curves.}
    \label{fig:hist_teaser}
\end{figure}

While efficient object detection is now an extensively researched problem in computer vision, this literature has mostly focused on the design of computationally efficient backbones. The introduction of heavy detection heads would reverse the computational gains that have been achieved with lightweight models~\cite{sandler2018mobilenetv2,ma2018shufflenet,han2020ghostnet}.
For example, the detection head of the Sparse RCNN~\cite{sun2021sparse} with $300$ proposals consumes $4\times$ the computation of the entire MobileNetV2 \cite{sandler2018mobilenetv2} ($25$ GFLOPS vs $5.5$ GFLOPS). In this work, we investigate whether it is possible to retain the accuracy gains and proposal sparsity of modern detection heads while reducing their computational cost, so as to make them applicable to efficient object detection design. We note that a main limitation of existing high-end detectors is that they treat all proposals equally, in the sense that the detection head applies to all proposals an operator of identical complexity, maintaining a constant cost per proposal. This, however, is unintuitive. While it seems appropriate to spend significant computation on good proposals, it is wasteful to allocate equal resources to poor proposals. Since the IoU of each proposal is known during training, the detector could, in principle, learn to allocate different amounts of computation to different proposals. This, however, requires a paradigm shift for detector design, illustrated in Figure \ref{fig:hist_teaser}: {\it that different proposals should be treated unequally in terms of resource allocation, reserving more computation for high quality proposals than low quality ones\/}.

The difficulty is that, because IoUs are not available at inference, the network has to {\it learn\/} to perform the resource allocation on the fly. This implies the need for a {\it resource allocation function\/} that depends on the proposal itself and has to be learned, i.e. a dynamic network module. To address this problem, we propose the {\it dynamic proposal processing\/} (DPP) framework, where the single operator used by current detection heads is replaced by an {\it operator set\/}, composed of multiple operators of different complexities. The benefit of this approach is to allow the detector to operate on multiple points of the complexity-precision curve, on a proposal by proposal basis, so as to optimize the overall trade-off between the two objectives. 
This is implemented by the addition of a {\it selection model\/} that chooses the best operator to apply to each proposal, at each stage of the network. We show that this selector can be very lightweight, a multi-layer perceptron that outputs a one-hot code over operator indices, and learned at training time, in an end-to-end manner. This is enabled by the introduction of two novel loss functions, which jointly encourage the allocation of the available computational budget to proposals of large IoU. An IoU loss teaches the detector to recognize proposals of large IoU and improve their alignment with ground truth bounding boxes. A complexity loss makes the selector aware of the number of instances, per image, so as to dynamically control the allocation of computational resources and meet the overall computational target.

Experimental results on the COCO dataset show that DPP achieves a better complexity-precision curve (see Figure \ref{fig:hist_teaser}) than designs that treat proposals equally, especially in the low complexity regime, confirming the effectiveness of treating proposals unequally. For large backbones (ResNet \cite{he2016deep}), DPP achieves the best precision-complexity curves in the literature, achieving state-of-the-art precision with $60\%$ of the computation of current models. For low-complexity networks (MobileNet \cite{sandler2018mobilenetv2}), the gains are even more significant, in that DPP establishes a new state-of-the-art in terms of both precision and computation and produces the best latency-precision curves.

\section{Related Work}

\label{sec:relw}
\noindent\textbf{Object detection.} Object detection frameworks can be mainly categorized into one-stage  \cite{liu2016ssd,lin2017focal,tian2019fcos,law2018cornernet,li2021generalized,zhang2020bridging} vs two-stage  \cite{ren2015faster,cai2018cascade,cai2016unified,dai2016r,chen2018context}, depending on the approach used to generate proposals. One-stage detectors can be anchor-based or not, but all rely on the very dense generation of proposals, which means each feature vector in the feature map is leveraged as a proposal. Two-stage detectors rely on a region proposal network \cite{ren2015faster} to filter out the majority of regions that are unlikely to contain an object instance. All the aforementioned methods require a post-processing step (non-maximum suppression) to remove a large number of duplicate proposals. More recently, an attention based framework \cite{carion2020end,zhu2020deformable,zhang2021detr,liu2021wb} has been proposed to overcome this problem, eliminating the need to post-process candidate predictions. By resorting to an attention mechanism, \cite{sun2021sparse} even showed that it is possible to rely on a very sparse proposal density. In result, existing methods differ significantly in terms of proposal density. However, within each framework, all proposals are treaty equally. In this paper, we show that, by diversifying the complexity of proposal processing dynamically, it is possible to reduce detection complexity without decreasing precision.

\noindent\textbf{Dynamic network. }Dynamic networks are a family of networks with input dependent structures or parameters derived from dynamic branches \cite{hu2018squeeze}. For classical convolutional networks, this can be done by using input-dependent rather than static filters~\cite{yang2019condconv,chen2020dynamic,li2021revisiting,li2021dynamic,verelst2020dynamic,li2021dynamicgating} or reweighing features spatially or in a channel-wise manner \cite{hu2018squeeze,hou2021coordinate}. Transformers are by definition dynamic networks, due to their extensive reliance on attention. Beyond that, \cite{rao2021dynamicvit,wang2021not} dynamically discard uninformative tokens to reduce computational cost. While previous methods show remarkable improvements in network efficiency, they mainly focus on backbones. This cannot fully address the problem of object detection, namely the heavy computation required to process proposals. Dynamic DETR \cite{dai2021dynamic} attempts to address the problem by building dynamic blocks on the detection head. However, it still processes all proposals with a common operator, inducing a constant complexity per proposal. In this work, we propose to leverage the power of dynamic networks by matching proposals to operators of variable complexity in a dynamic manner.

\section{Complexity and Precision of Proposals}
\label{sec:method}

In this section, we compare the complexity of treating proposals equally or unequally. We assume that a backbone produces a set $\bm{X}=\{\bm{x}_1, \bm{x}_2,...,\bm{x}_N\}$ of proposals and focus on the cost of the detection head, i.e. ignore backbone costs. We further assume that the computation of the detection head can be decomposed into a per-proposal operator $h$, e.g. a network block, and a pairwise component $p$ that accounts for the cost of inter-proposal computations. For example, the NMS operation of classical detectors or a self-attention mechanism between proposals for transformers. 

\noindent\textbf{Complexity of equally treated proposals.} In prior works, all proposals are processed by the same operator $h$.
This has complexity
\begin{equation}
    \mathcal{C}(\psi)= NC_h + \frac{N(N-1)}{2}C_p,
    \label{eq:static_pro}
\end{equation}
where $\psi=\{h,p\}$, and $C_h$ and $C_p$ are the per proposal complexity of $h$ and $p$, respectively.

\noindent\textbf{Complexity of unequally treated proposals.}
We propose to treat proposals unequally. Rather than applying the same operator $h$ to all proposals, we propose to leverage an operator set $\mathcal{G} = \{h_j\}_{j=1}^J$ of $J$ operators of different architectures and complexity, which are assigned to the proposals $\bm{x}_i$ by a dynamic selector $s$. This has complexity
\begin{align}
    \mathcal{C}(\psi) = \sum_{i=1}^NC_{h_{s_i}} + \frac{N(N-1)}{2}C_p,
    \label{eq:dy_pro}
\end{align}
where $s_i=s(\bm{x}_i)$, $h_{s_i} \in\mathcal{G}$ represents the operator from $\mathcal{G}$ that is assigned to the proposal $\bm{x}_i$ by the selector $s$, $\psi=\{\{h_{s_i}\}_i,s,p\}$, and $C_{h_{s_i}}$ is the complexity of the entire per proposal operation (selector plus operator). For simplicity, the pairwise complexity is still considered constant. 

\noindent\textbf{Precision over proposals.} When the detection head treats proposals unequally, the optimal detector precision for a given complexity constraint $C$ can be determined by optimizing the assignment of operators to proposals
\begin{align}
    P(\psi^*| C) = \max_{\underset{\mathcal{C}(\psi)<C}{h_{s_i}\in\mathcal{G}}}\mathcal{P}(\{h_{s_i}\}_i), \label{eq:opt}
\end{align}
where $\mathcal{P}(\{h_{s_i}\}_i)$ is the precision of a specific operator assignment $\{h_{s_i}\}_i$. As $C$ changes, $P(\psi^*|C)$ forms a complexity-precision (C-P) curve that characterizes the optimal performance, in terms of the trade-off between cost and precision, of the object detectors implementable with $\mathcal{G}$. In this work, we use both precision (mAP) and the C-P curve as criteria to justify the effectiveness of treating proposals unequally. Note that the assignment of operators to proposals is the key to optimize the precision under a given computation budget $C$. This is formulated as a learning function implementable with a simple network branch and solved via suitable loss functions, as discussed next.

\section{Dynamic Proposal Processing}
In this section, we proposed a \textit{dynamic proposal processing} (DPP) framework for the solution of~(\ref{eq:opt}).
Following the design of prior works \cite{carion2020end,zhu2020deformable,sun2021sparse}, we assume a detector head composed of multiple stages ($\psi=\phi_1 \circ \ldots \circ \phi_K$) that process proposals sequentially. Each stage $\phi_k$ is implemented with an operator chosen from $\mathcal{G}$ by a selector $s$. To minimize complexity, the selector can be applied only to a subset  $k\in\mathcal{K} \subset \{1, \ldots, K\}$ of the stages, with the remaining stages using the operator chosen for their predecessor, i.e. $ \phi_k = \phi_{k-1}, \forall k \not\in\mathcal{K}$.

\subsection{Operator Set}
In this paper, we consider an operator set $\mathcal{G}=\{g_0, g_1, g_2\}$ composed of three operators of very different computational cost. Specifically, $g_0$ is a high complexity operator, implemented with a dynamic convolutional layer (DyConv) of proposal dependent parameters and a feed forward network (FFN) \cite{carion2020end}. This operator is based on the dynamic head architecture employed in the recent Sparse R-CNN \cite{sun2021sparse}. $g_1$ is a medium complexity operator, implemented with a static FFN \cite{carion2020end}. Finally, $g_2$ is a light operator formed by an identity block, which simply feeds the proposal forward with no further refinement.

\subsection{Selector}
\begin{figure}[t]
\centering
\includegraphics[width=0.58\linewidth]{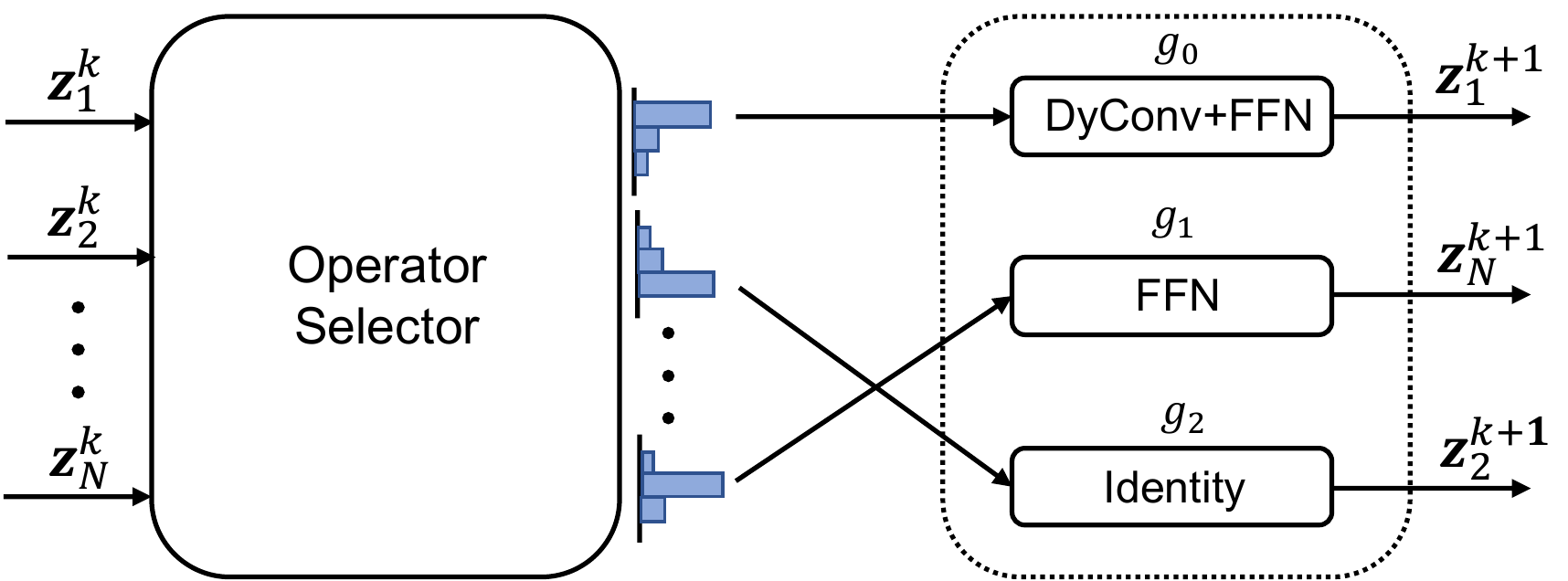}
\caption{\textbf{Flow graph of operator assignments to proposals}. The selector takes the proposal embeddings, i.e. $\{\bm{z}_1^k$, $\bm{z}_2^k$,..., $\bm{z}_N^k\}$ in the $k^{th}$ stage as input and outputs a selection vector per proposal. The operator that matches the index of largest value in the selection vector is selected to process the proposal. In the operator set, operator $g_0$ contains a high complexity dynamic convolution (DyConv) followed by a FFN \cite{carion2020end}. $g_1$ consists of a feed forward network (FFN), while $g_2$ is implemented with an identity function (Identity).}
\label{fig:flow}
\end{figure}

In DPP, the selector is the key component to control the trade-off between precision and complexity, by controlling the assignment of operators to proposals. 
Let $\bm{z}^k_i$ be the embedding of proposal $\bm{x}_i$ at the input of stage $\phi_k$. The selector is implemented with a $3$-layer MLP that associates a $3$ dimensional vector $\epsilon^k_i \in [0,1]^3$ with $\bm{z}^k_i$ according to
\begin{align}
    \epsilon^k_i = \textrm{MLP}(\bm{z}^k_i) \label{eq:selector}
\end{align}
where $\epsilon^k_{i,j}$ is the selection variable in $\epsilon^k_i$ that represents the strength of the assignment of operator $g_j$ to proposal $\bm{x}_i$. 
During training, the selection vector is a one hot code over three variables and the Gumble-Softmax function \cite{jang2016categorical} is used as activation of the MLP to generate the selection vector. For inference, the selection variables have soft values and the operator that matches the index of the selection variable with largest value is chosen. The flow graph of the operator assignment  process is illustrated in Figure~\ref{fig:flow}. Please note that the proposed selector is very light (using 4e-3 GFLOPS for $100$ proposals in our experimental setting), in fact negligible in complexity when compared to the detection head.

It is clear from (\ref{eq:selector}) that the chosen operator varies both across proposals $i$ and head stages $k$, enabling the unequal treatment of proposals in a dynamic manner. Furthermore, while ${\cal G}$ has cardinality three, the cardinality of the set of network architectures that can be used to implement the detector head is $3^{|{\cal K}|}$. Finally, because the selector is trainable, the assignment function can be learned end-to-end. 

\subsection{Loss Functions}
To assure that, given a complexity budget, DPP selects the optimal sequence of operators for each proposal, a selection loss is applied to the selector of each stage in ${\cal K}$. This selection loss is designed to encourage two goals. First, complex operators should be assigned to high quality proposals (large IoU), since these require most additional work by the detection head. 
This is enforced through the IoU loss
\begin{align}
    L_{iou} = \frac{1}{N}\sum_{i=1}^N \sum_{k \in {\cal K}} \sum_{j\in\{0,1\}} (1-u_i^k)\epsilon^k_{i,j} + u_i^k(1-\epsilon^k_{i,j}), \label{eq:iou}
\end{align}
where $u_i^k$ is the IoU of the $i^{th}$ proposal in $k^{th}$ stage. $L_{iou}$ pushes the selector to turn $\epsilon_{i,0}^k$ and $\epsilon_{i,1}^k$ into `0' for proposals of IoU smaller than 0.5 and into `1' otherwise. This encourages the use of more complex operators in stage $k$ for the high quality proposals, which require more efforts for classifying categories and regressing bounding boxes. Moreover, the loss magnitude is determined by the IoU value, originating larger gradients when the selector predicts $\epsilon_{i,0}^k$ or $\epsilon_{i,1}^k$ as `1' for tiny IoU proposals or as `0' for large proposals. 
Second, the selector should be aware of the total number of instances in each image and adjust the overall complexity according to it, i.e., selecting more complex operators when instances are dense. This is enforced through the complexity loss
\begin{align}
    L_c=\frac{1}{N}\sum_{k \in {\cal K}}\left|\sum_{i=1}^N \epsilon^k_{i,0}-T\right|,\label{eq:loss_n}
\end{align}
where $T$ is the target number of times that operator $g_0$ is selected for a particular image. This is defined as $T=\alpha M$ where $\alpha$ is a multiplier that specifies a multiple of the $M$ object instances in the image. Moreover, the condition $T \in[T_{min}, N]$ is enforced, by clipping $\alpha M$ according to a pre-specified lower bound $T_{min}$ and an upper bound given by the overall proposal number $N$. The lower bound prevents a very sparse selection of high complexity operators $g_0$ and $\alpha$ then adjusts the selector according to the number of instances. $\alpha$, $T_{min}$ and $N$ are hyperparameters that can be leveraged to modify the behavior of DPP, as discussed in the experimental section.

The overall selection loss is finally 
\begin{align}
    L_{s} = L_{iou}+\lambda L_{c},
\end{align}
where $\lambda$ is the hyperparameter that controls the trade-off between the loss components. Note that the selection loss is a plug-and-play loss that can be applied to different object detectors. In this paper, $L_{s}$ is combined with all the losses of the original detector to which DPP is applied, including the cross entropy loss and the bounding box regression loss, which are omitted from our discussion. 

\section{Experiments}
\noindent{\textbf{Dataset.}} DPP is evaluated on the COCO dataset \cite{lin2014microsoft}. It is trained on the train2017 split and mainly tested on the val2017 split with mAP.

\noindent{\textbf{Network.}} DPP is applied to detectors whose backbone is built on MobileNet V2 \cite{sandler2018mobilenetv2} or ResNet-50 \cite{he2016deep}, using Feature Pyramid Networks (FPN) \cite{lin2017feature}, on top of which proposals are generated using the strategies of \cite{sun2021sparse}. For MobileNetV2 \cite{sandler2018mobilenetv2}, the FPN only considers features with strides $16$ and $32$ and the $3\times3$ FPN convolution is decomposed into an $1\times1$ pointwise convolution and a $3\times3$ depthwise convolution for efficiency. For ResNet-50, FPN is implemented on features with the standard $4$ different strides. Following \cite{sun2021sparse}, the detection head is a decoder only transformer of $6$ stages. For simplicity, the selector is only applied in stages $\mathcal{K} = \{2,4,6\}$. In the first stage, all proposals are processed with the high complexity operator ($g_0$). The full operator set $\cal G$ is used in all remaining stages.

\noindent{\textbf{Experimental setting.}} DPP is pretrained without selectors, using the hyperparameters and data augmentations of \cite{carion2020end,sun2021sparse,ren2015faster} on COCO. The selectors are then added and trained with learning rate $2$e-$5$, while $2$e-$6$ is used for other layers. The training process lasts $36$ ($3\times$) epochs and the learning rate is divided by $10$ at $27$ and $33$ epochs. The selection loss $L_s$ is combined with all the losses used in \cite{carion2020end,sun2021sparse,ren2015faster}, i.e. cross entropy loss $L_{ce}$, GIoU loss $L_{giou}$ and bounding boxes   regression loss $L_{bbox}$ and $\lambda=10$ is used for all selectors. The lower bound $T_{min}$ for the target number of times that $g_0$ is selected is manually set to $T_{min}^{last}$ for the last selector. For the remaining selectors, $T_{min}$ is derived automatically, so that it decreases exponentially from $N$ (number of proposals) to $T_{min}^{last}$. Hence we omit the subscript of the lower bound for conciseness in the following sections. The multiplier $\alpha$ is constant across all selectors.

\subsection{Proposal processing by DPP}
We start by discussing experiments that illustrate the unequal processing of proposals by DPP and how this impacts complexity. In these experiments, we analyze how each operator contributes to the processing of proposals produced by a ResNet-50 backbone. DPP training uses a lower bound $T_{min}=1$, a multiplier $\alpha=2$, and $N=100$ proposals.

\noindent\textbf{Contribution of Each Operator.} The influence of each operator in $\mathcal{G}=\{g_0, g_1, g_2\}$ is investigated separately. For this, we manually split the proposals into three groups,  according to the operator that process them, and evaluate precision for each group. For simplicity, the split is only based on the selector of the last DPP stage, i.e. the analysis is limited to this stage.

\begin{table}[t!]
	\begin{center}
	    \small
	    \setlength{\tabcolsep}{4.7mm}{
		\begin{tabular}{@{\hskip 7mm}c@{\hskip 10.3mm}c@{\hskip 10.3mm}c@{\hskip 7mm}|@{\hskip 7mm}c@{\hskip 10.3mm}c@{\hskip 10.3mm}c@{\hskip 7mm}|@{\hskip 7mm}c@{\hskip 7mm}}
		    \specialrule{.1em}{.05em}{.05em}
		     $g_0$ & $g_1$ & $g_2$ & AP & AP$_{50}$ & AP$_{75}$ & $N_{eval}$\\
		    \hline
		    \checkmark  &  & & 41.0&58.5&44.4 & 15\\
		     &\checkmark & &2.4&5.1&2.1 & 7\\
             & & \checkmark& 0.8&2.2&0.5 &78\\
             \checkmark & \checkmark & & 41.7&59.9&45.2 &22\\
             \checkmark & \checkmark & \checkmark & \textbf{42.2}&\textbf{60.6}&\textbf{45.5} &100\\
	       \specialrule{.1em}{.05em}{.05em}
		\end{tabular}
		}
	\end{center}
	\caption{\textbf{Contribution of each operator to proposal processing.} Performance is evaluated on the COCO validation set. $N_{eval}$ is the average number of proposals matched to the checkmarked operator(s).}
	\label{table:abl_operator}
\end{table}
Table \ref{table:abl_operator} shows the precision of proposals processed by the different operators. $N_{eval}$ represents the average number of proposals evaluated across the COCO validation set. This is equivalent to the number of times that the operators checkmarked in the table were used. Clearly, the proposals processed by $g_0$ are the main contributors to the overall precision ($41.0$ vs $42.2$), even though only $15$ such proposals are evaluated on average. For proposals processed by $g_1$ or $g_2$ the performance is quite poor. This shows that the selector successfully allocates operators to proposals, assigning the operators of large complexity to the proposals that have higher chance of being associated with objects and devoting much less computation to the remaining.

Interestingly, the vast majority of proposals ($78\%$) are assigned to $g_2$, i.e. use {\it no\/} computation  in the final DPP stage. These are very poor proposals (almost zero AP), showing that the DPP detector learns to ``give up'' on such proposals, simply shipping them to the output. When the proposals processed by $g_0$ and $g_1$ are merged, the precision is promoted by $0.7\%$ ($41.0$ vs $41.7$). This shows, that the two types of proposals are complementary and confirms that $g_1$ is important although sparsely used.

\begin{figure}[t]
\centering
\includegraphics[width=0.75\linewidth]{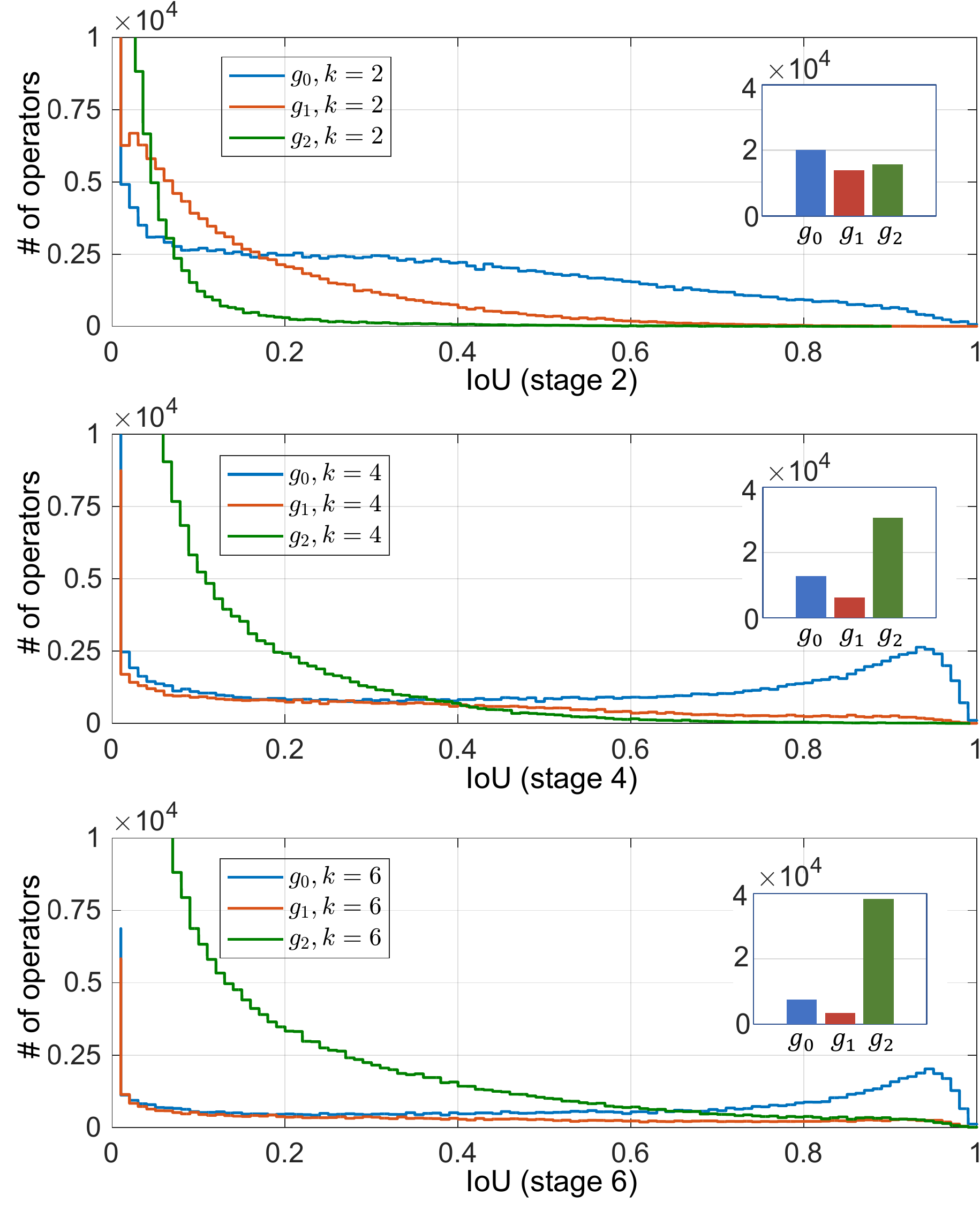}
\caption{\textbf{IoU distribution of proposals matched to the three operators across DPP stages} (stage indexes $k\in\{2,4,6\}$). Within each plot, the number of proposals processed per operator is shown as a subplot. }
\label{fig:iou_hist}
\end{figure}

\noindent\textbf{Performance of Each Stage in DPP.} Precision is tested across all stages ($k=1\sim6$) and we obtain the AP as $\{15.6,32.1,39.3,41.7,42.0, 42.2\}$. The results show that the precision increases quickly in the first $4$ stages and then saturates. Among the $6$ stages, the selector is applied in stages $\{2,4,6\}$. 
The IoU distribution of the proposals selected by different operators is shown in Figure \ref{fig:iou_hist}. The total number of proposals processed by each operator is illustrated as a subplot. Note that the proposals of larger IoU are mostly processed by $g_0$ (blue curve) even though the overall number of proposals processed by $g_0$ decreases drastically for the later stages (blue  bar). In these stages, most proposals are simply ``shipped to the next stage'' without any computation ($g_2$). Conversely, most low IoU proposals are processed by operator $g_2$ (green curve) and the number of such proposals increases drastically with the stage (green bar). This illustrates how DPP is quite successful at trading off complexity for precision.  

\begin{figure}[t]
\centering
\includegraphics[width=0.95\linewidth]{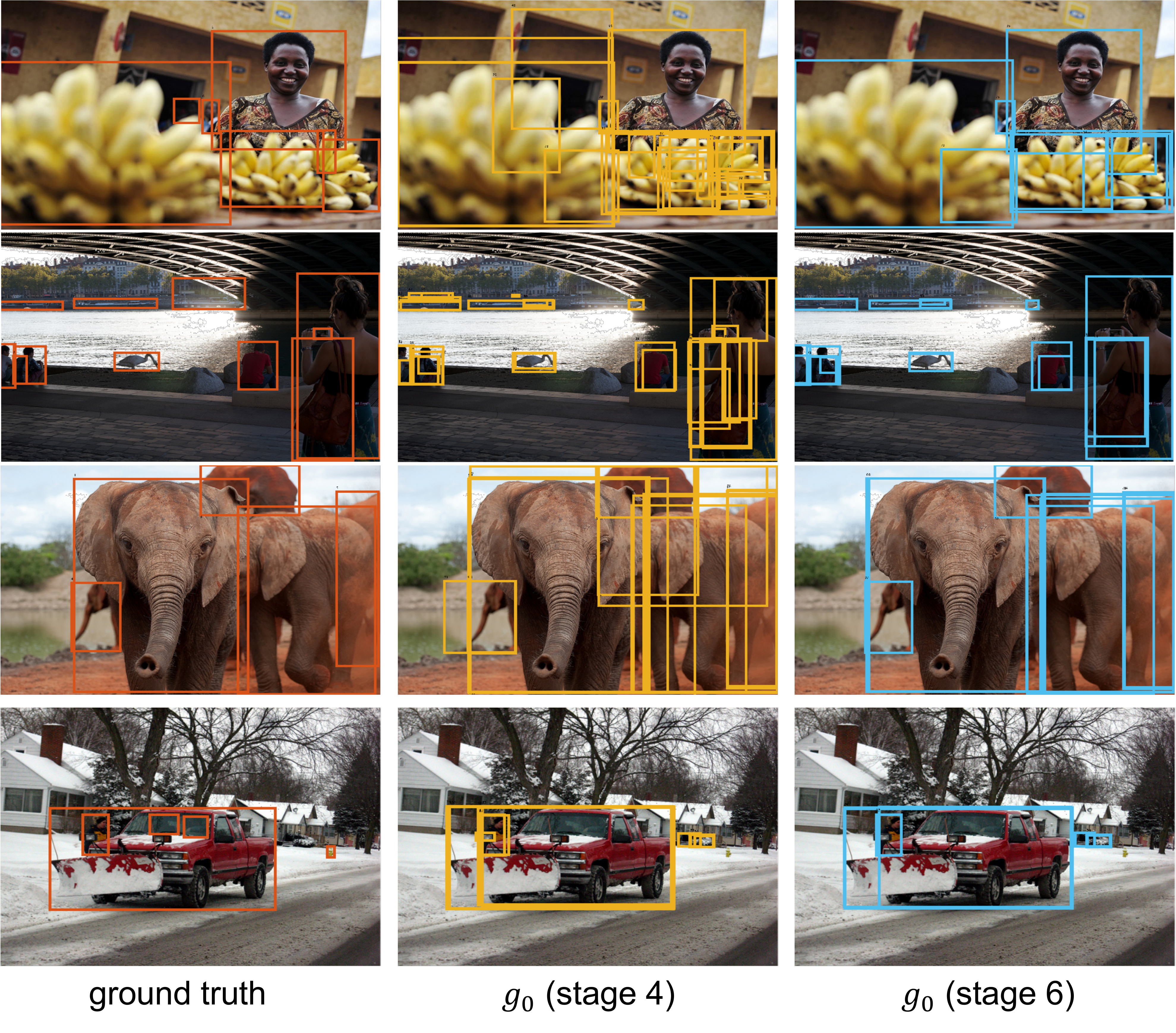}
\vspace{-0.5em}
\caption{Boxes predicted by operator $g_0$ in stages $4$ and $6$. }
\label{fig:vis}
\end{figure}

\noindent\textbf{Visualization.}
Figure \ref{fig:vis} shows some qualitative results, in the form of bounding boxes predicted by the high complexity operator $g_0$ in stages $4$ and $6$. Note that the boxes predicted in stage $6$ (right column) have a good overlap with the ground truth (left column) with limited duplication. By comparing the boxes predicted in different stages, we can observe that they are refined in the deeper stages. More importantly, duplication is removed to a remarkable extent, indicating that not only the selector  prevents poor proposals from being processed by operator $g_0$ but the network gradually transforms duplicates into bad proposals, in order to meet the complexity constraint. 

\subsection{Main Results}
DPP is compared to the state-of-the-arts for two backbones, ResNet-50 \cite{he2016deep} and MobileNetV2 \cite{sandler2018mobilenetv2}, with the results shown in Table \ref{table:main_res} and \ref{table:main_mv2} respectively. In Table \ref{table:main_res} and \ref{table:main_mv2}, $\Bar{N}$ represents the number of proposals for the Faster R-CNN \cite{ren2015faster} and Sparse R-CNN \cite{sun2021sparse} and the number of queries (which play identical roles to proposals for final prediction) for attention baselines. For DPP, where proposals are processed by different operators, $\Bar{N}$ is the equivalent proposal number, defined as the ratio between the overall FLOPS spent by the detector head and the FLOPS spent by the high complexity operator $g_0$. 
For ResNet-50 the results of the baselines are copied from the original papers. For MobileNetV2 baselines, they are obtained with the official code, using the recommended hyperparameters. 

\begin{figure}
\centering
\begin{minipage}{.48\textwidth}
  \centering
  \includegraphics[width=.98\linewidth]{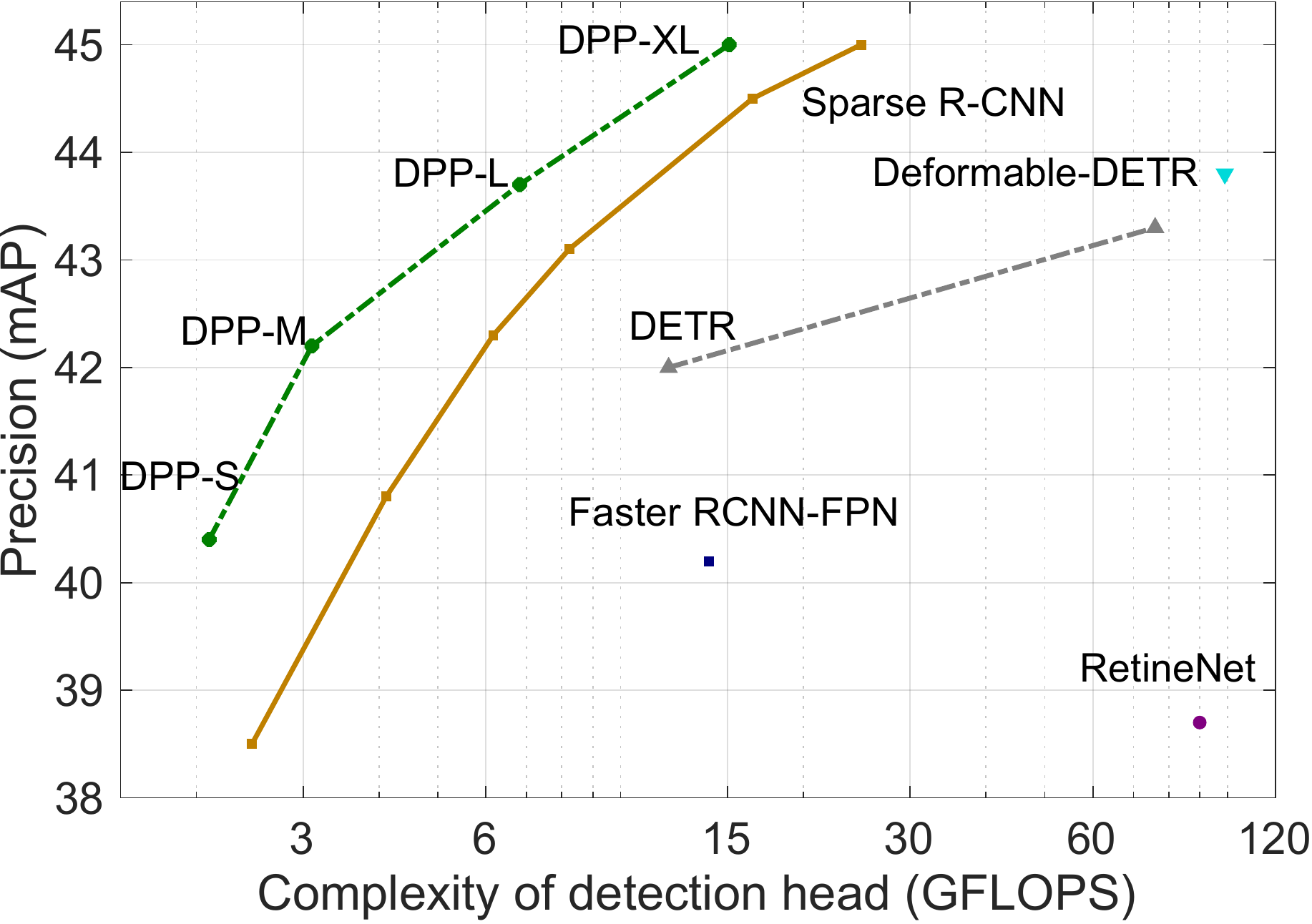}
  \captionof{figure}{\textbf{Comparison of complexity-precision curve on ResNet-50} between DPP and state-of-the-arts (MAdds only reflects the computational cost of the detector head).}
  \label{fig:res_pc_funct}
\end{minipage}\quad
\begin{minipage}{.48\textwidth}
  \centering
  \includegraphics[width=.98\linewidth]{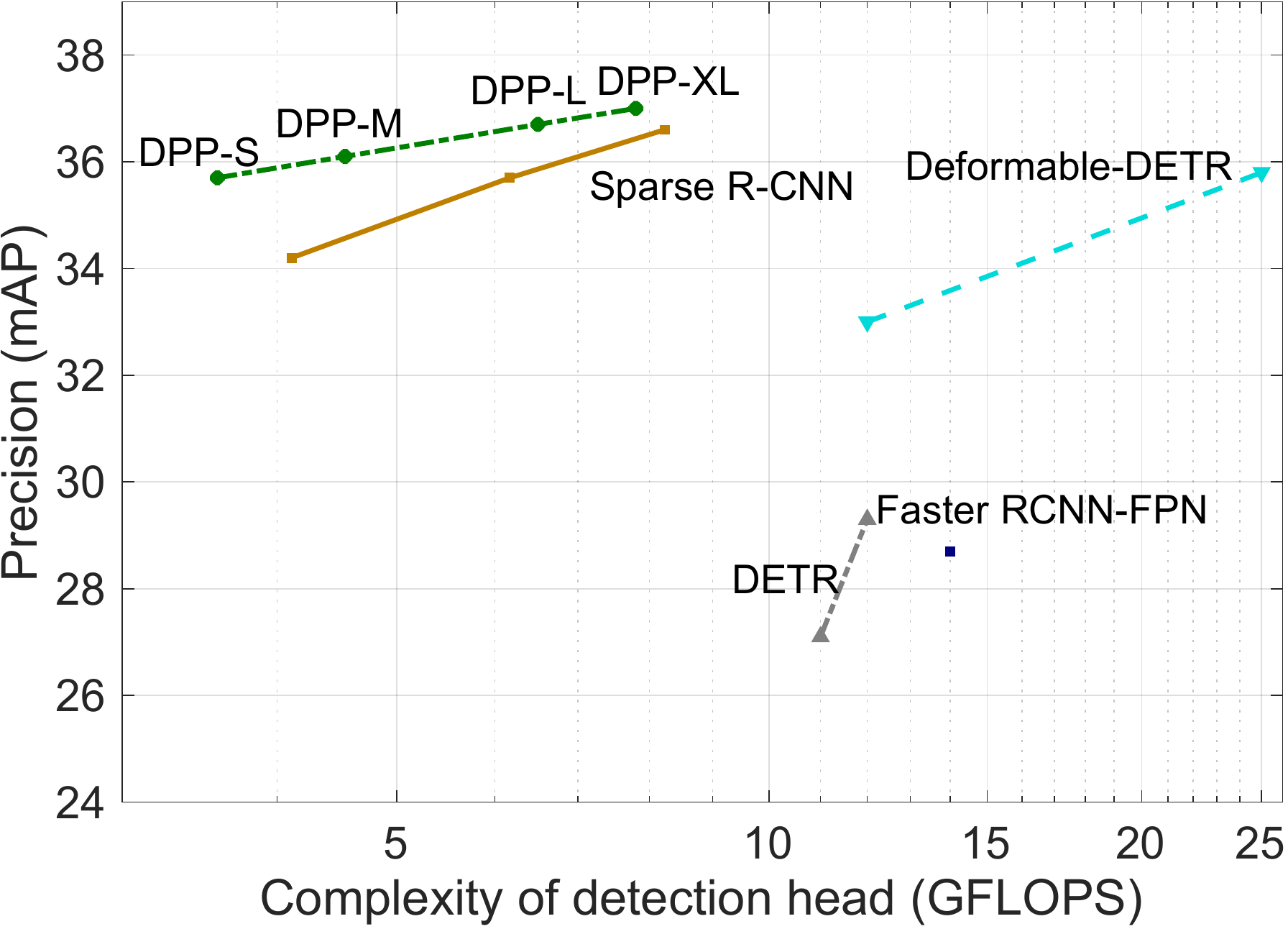}
  \captionof{figure}{\textbf{Comparison of complexity-precision curve on MobileNetV2} between DPP and state-of-the-arts (MAdds only reflects the computational cost of the detector head).}
  \label{fig:mv2_pc_funct}
\end{minipage}
\end{figure}

\noindent\textbf{ResNet.} When ResNet-50 is used as backbone, four variants of DPP are used as the detection head. DPP-S, DPP-M and DPP-L use different overall numbers of proposals ($N\in\{50,100,300\}$). The other hyperparameters, i.e. $T_{min}$ and $\alpha$, are set to $1$ and $2$ respectively. In this way, $L_c$ can assure there is at least $1$ high complexity operator and assign $2$ high complexity operators per instance, on average. DPP-XL, is equivalent to DPP-L but further increases the hyperparameter $T_{min}$ to $100$. Table \ref{table:main_res} shows that DPP achieves a good trade-off between complexity and precision. At the high end, DPP-XL performs on par with the Sparse R-CNN \cite{sun2021sparse} with a much lighter detection head ($15$ vs $25$ GFLOPS). The prior method with this level of complexity (Faster RCNN-FPN) has an AP loss of close to $5$ points ($40.2\%$ vs $45.0\%$). At the low end, DPP-S reduces the Sparse R-CNN computation by $12.5\times$, for a decrease of $4.6$ points in AP. This is equivalent to the Faster RCNN-FPN, but saving $7\times$ computation. Figure \ref{fig:res_pc_funct} shows that the complexity-precision (C-P) curve of DPP is better than those of all other baselines. This confirms the benefits of treating proposals unequally. Finally, DPP is evaluated on the COCO test set and we achieve the AP as $\{44.7, 43.8, 42.5, 40.7\}$ for the four variants of DPP ($44.7$ is obtained for SparseRCNN \cite{sun2021sparse}), which further justifies the effectiveness and stability of DPP.

\begin{table}[t!]
	\begin{center}
	    \small
	    \setlength{\tabcolsep}{1mm}{
		\begin{tabular}{l@{\hskip 1.1mm}|@{\hskip 1.1mm}c@{\hskip 1.1mm}|@{\hskip 1.1mm}c@{\hskip 1.1mm}|c@{\hskip 1.1mm}c@{\hskip 1.1mm}c@{\hskip 1.1mm}c@{\hskip 1.1mm}c@{\hskip 1.1mm}c@{\hskip 1.1mm}|c@{\hskip 1.1mm}}
		    \specialrule{.1em}{.05em}{.05em}
		    Method & $\Bar{N}$ & Epochs & AP & AP$_{50}$ & AP$_{75}$ & AP$_{s}$ & AP$_{m}$ & AP$_{l}$ & GFLOPS \\
		    \hline
		    Faster RCNN-FPN \cite{ren2015faster}& 2000 & 36 & 40.2 & 43.8 & 43.8 & 24.2 & 43.5 & 52.0 & 14\\
		    RetinaNet \cite{lin2017focal}& - & 36 &38.7 & 58.0 & 41.5 & 23.3 & 42.3 & 50.3 & 90\\
            DETR-DC5 \cite{carion2020end}& 100 & 500 &43.3 & 63.1 & 45.9 & 22.5 & 47.3 & \textbf{61.1} & 76 \\
            Deformable Detr \cite{zhu2020deformable} & 300 & 50 & 43.8 & 62.6& 44.2 & 20.5& 47.1 &58.0&98\\
            Sparse R-CNN \cite{sun2021sparse}& 300 & 36 & \textbf{45.0}&\textbf{64.1}&\textbf{49.0}&27.8&\textbf{47.6}&59.7 & 25\\
            \hline
            DPP-XL (ours)& 182.3 &36 & \textbf{45.0}&63.8&48.8&\textbf{28.2}&47.4 & 59.9 & 15\\
            DPP-L (ours)& 82.6 &36 & 43.7&62.4&47.5&27.2&46.0&59.1 & 6.8\\
            DPP-M (ours)& 38.8 &36 & 42.2&60.6&45.5&23.9&44.6&58.5 & 3.2\\
            DPP-S (ours)& 25.5 &36 & 40.4&58.2&43.4&22.0&42.8&57.0 & 2.1\\
	       \specialrule{.1em}{.05em}{.05em}
		\end{tabular}
		}
	\end{center}
	\caption{\textbf{Comparison to state-of-the-art object detectors on COCO validation set with ResNet-50}. Four variants of DPP with various sizes are shown, based on FPN. For DPP, $\Bar{N}$ is the equivalent proposal number, defined as the ratio between the overall FLOPS spent by the detector head and the FLOPS spent by each high complexity operator $g_0$ ($\Bar{N}=C(\psi)/C_{g_0}$ in (\ref{eq:dy_pro})), while for baselines $\Bar{N}$ is either the proposal number or the number of queries. The complexity (GFLOPS) is only that of the the detection head.}
	\label{table:main_res}
\end{table}

\noindent\textbf{MobileNetV2.} For MobileNetV2 we consider a lighter detection head, by decreasing the number of proposals for both DPP and baselines. Similar to ResNet, four variants of DPP are proposed. Given the more important role played by the detection head in this case, we force the selector to choose more high complexity operators, by increasing the lower bound $T_{min}$ for the target number of the operators $g_0$. When comparing DPP to the state-of-the-arts, the results shown in Table \ref{table:main_mv2} and Figure \ref{fig:mv2_pc_funct} enable even stronger conclusions than those drawn for the ResNet. In this case, DPP-XL outperforms the Sparse R-CNN, establishing a new state of the art, and even DPP-S has a small AP loss (less than $1\%$) compared to the latter.  These results confirm that DPP is a generic framework, which can perform well with different types of backbones. 
\vspace{-1.0em}

\begin{table}[t!]
	\begin{center}
	    \small
	    \setlength{\tabcolsep}{1mm}{
		\begin{tabular}{l@{\hskip 1.1mm}|@{\hskip 1.1mm}c@{\hskip 1.1mm}|@{\hskip 1.1mm}c@{\hskip 1.1mm}|c@{\hskip 1.1mm}c@{\hskip 1.1mm}c@{\hskip 1.1mm}c@{\hskip 1.1mm}c@{\hskip 1.1mm}c@{\hskip 1.1mm}|c@{\hskip 1.1mm}}
		    \specialrule{.1em}{.05em}{.05em}
		    Method & $\Bar{N}$ & Epochs & AP & AP$_{50}$ & AP$_{75}$ & AP$_{s}$ & AP$_{m}$ & AP$_{l}$ & GFLOPS \\
		    \hline
		    Faster RCNN-FPN \cite{ren2015faster}& 2000 & 36 & 28.7&47.1&30.3&12.7&32.6&39.6 & 14\\
            DETR \cite{carion2020end}& 100 & 150& 29.3&49.0&29.1&9.8&30.9&47.5 &12 \\
            Deformable Detr \cite{zhu2020deformable}& 100 & 50 & 35.8&54.4&37.8&17.6&38.8&51.0 & 25\\
            Sparse R-CNN \cite{sun2021sparse}& 100 & 36 & 36.6&55.3&39.1&18.0&39.3 &\textbf{52.9}& 8.2\\
            \hline
            DPP-XL (ours)& 94.8 &36 & \textbf{36.9}&\textbf{55.8}&\textbf{39.3}&\textbf{18.8}&\textbf{40.3}&51.9 & 7.8\\
            DPP-L (ours)& 78.9 &36 & 36.7&55.3&39.0&18.4&39.8&52.2 & 6.5\\
            DPP-M (ours)& 54.5 & 36& 36.1&54.5&38.3&17.5&38.8&52.1 &4.5\\
            DPP-S (ours)& 43.7 & 36&35.7&54.2&37.9&16.9&38.2&52.1 &3.6 \\
            \specialrule{.1em}{.05em}{.05em}
		\end{tabular}
		}
	\end{center}
	\caption{\textbf{Comparison to state-of-the-art object detectors on COCO validation set with MobileNetV2}. Four variants of DPP with various sizes are shown, based on light FPN for features with $2$ strides ($16$, $32$). For DPP, $\Bar{N}$ is the equivalent proposal number, defined same as that in Table \ref{table:main_res}, while for baselines $\Bar{N}$ is either the proposal number or the number of queries. The complexity (GFLOPS) is only that of the the detection head.}
	\label{table:main_mv2}
\end{table}

\begin{figure}
\centering
\begin{minipage}{.48\textwidth}
  \centering
  \includegraphics[width=.96\linewidth]{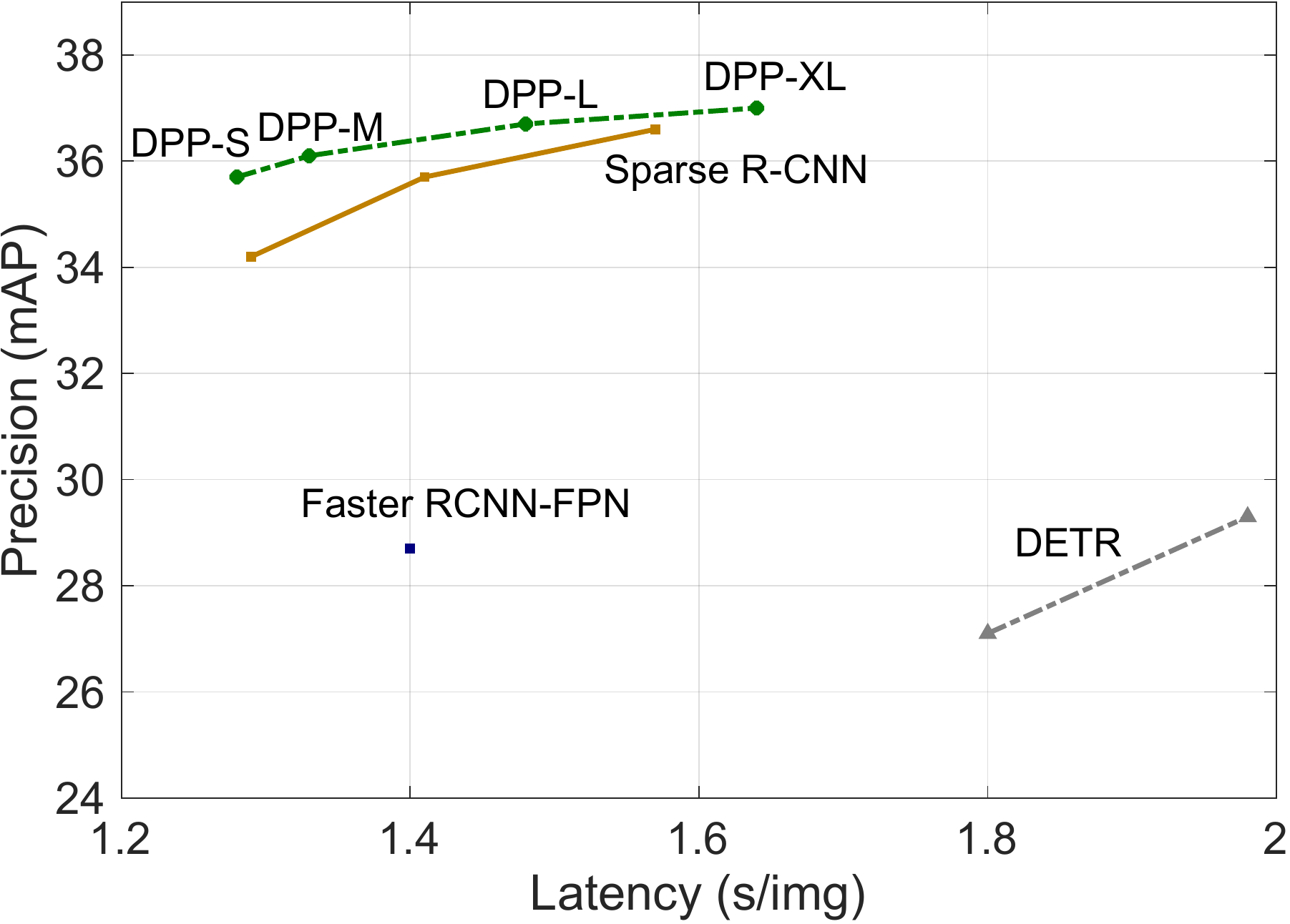}
  \captionof{figure}{\textbf{Comparison of latency-precision curve.} Latency is tested by using MobileNetV2 implemented on a CPU, across the COCO validation set. Latency reflects the inference time of the whole network instead of only the detection head.}
  \label{fig:mv2_latency}
\end{minipage}\quad
\begin{minipage}{.48\textwidth}
  \centering
  \includegraphics[width=1.0\linewidth]{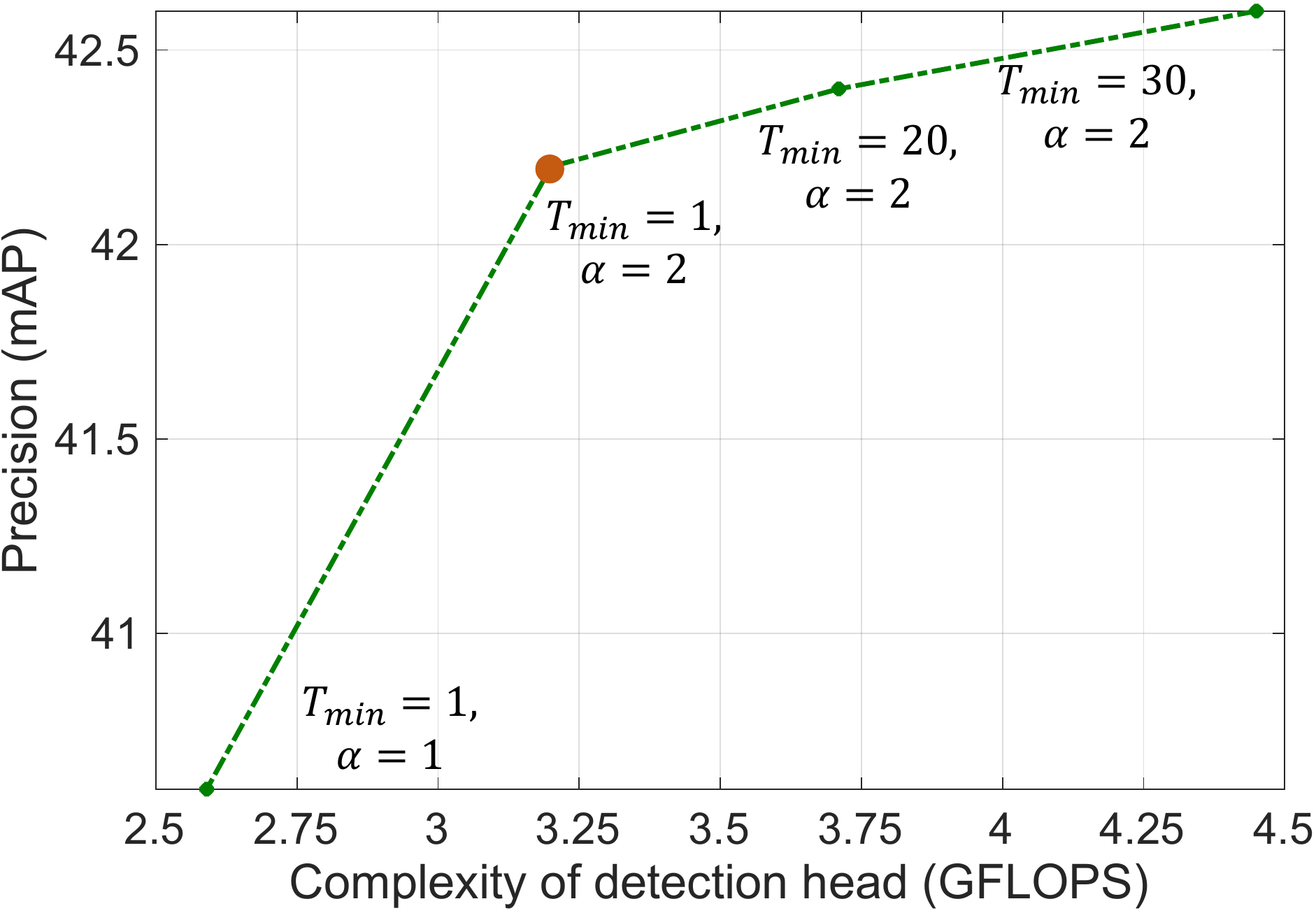}
  \captionof{figure}{\textbf{Effect of the hyperparameters for the target usage of $g_0$,} i.e. lower bound ($T_{min}$) and multiplier ($\alpha$), on the complexity-precision trade-off. The performance of four variants of DPP model are illustrated via varying $T_{min}$ and $\alpha$.}
  \label{fig:ddp_abl_dy}
\end{minipage}
\end{figure}
\vspace{-1.0em}
\noindent\textbf{Inference speed.} Inference speed is measured for DPP and baselines on MobileNetV2 with a single-threaded core Intel(R) Xeon(R) CPU E5-2470 (2.4GHz) (Deformable DETR \cite{zhu2020deformable} does not support CPU implementation). Results are obtained by averaging inference time of all images in the COCO validation split and shown in Figure \ref{fig:mv2_latency}. The latency is for the whole network, not just the head. It can be seen that DPP achieves a consistently better latency-precision curve and its savings in computation are clearly reflected in savings of inference time. 
\subsection{Ablation Study}
In this section, we present some ablation studies for the proposed loss function and hyperparameters used by DPP. The backbone is based on ResNet-50 and, by default, $100$ proposals ($N=100$) are used by all models. The lower bound $T_{min}$ and multiplier $\alpha$ for the operator $g_0$ are $1$ and $2$. All experiments are performed on the COCO dataset.

\noindent\textbf{Selection loss. }We start by exploring the influence of the selection loss $L_s$ on DPP performance.  Table \ref{table:abl_subloss} shows that using either component, IoU loss $L_{iou}$ or complexity loss $L_c$, alone degrades the precision of DPP. Without $L_{iou}$ the precision drops by $0.4\%$ ($41.8\%$ vs $42.2\%$), because the selector can no longer fully match the proposal qualities to the operator complexities. Without $L_{c}$, the precision is $1.1\%$ worse ($41.1\%$ vs $42.2\%$). This is because, during training without $L_{c}$, the model is more prone to assign the light operator ($g_2$) to proposals. Beyond weakening precision, this more critically prevents the complexity of DPP from being modified as needed. Table \ref{table:abl_lambda} studies the trade-off between $L_{iou}$ and $L_c$ as a function of $\lambda$.  The performance is similar for $\lambda=1$ and $\lambda=10$, but further increasing $\lambda$ makes DPP focus too much on complexity and ignore the importance of IoU matching, degrading performance. We thus set $\lambda=10$ in all subsequent experiments.

\begin{table}[t!]
	\begin{center}
	    \small
	    \setlength{\tabcolsep}{2.7mm}{
		\begin{tabular}{c@{\hskip 4.5mm}c|@{\hskip 4.5mm}c@{\hskip 9mm}c@{\hskip 9mm}c@{\hskip 9mm}c@{\hskip 9mm}c@{\hskip 9mm}c@{\hskip 9mm}c}
		    \specialrule{.1em}{.05em}{.05em}
		    $L_{iou}$ & $L_c$ & AP & AP$_{50}$ & AP$_{75}$ & AP$_{s}$ & AP$_{m}$ & AP$_{l}$\\
		    \hline
		    \checkmark& &41.1&59.2&44.2&23.1&43.1&57.0\\
		      & \checkmark & 41.8&60.2&45.1&23.7&44.0&58.2\\
             \checkmark & \checkmark & \textbf{42.2}&\textbf{60.6}&\textbf{45.5} &\textbf{23.9}&\textbf{44.6}&\textbf{58.5}\\
	       \specialrule{.1em}{.05em}{.05em}
		\end{tabular}
		}
	\end{center}
	\caption{\textbf{Effect of the loss functions}, i.e. the IoU loss $L_{iou}$ and the complexity loss $L_c$ on DPP.}
	\label{table:abl_subloss}
\end{table}
\begin{table}[t!]
	\begin{center}
	    \small
	    \setlength{\tabcolsep}{2.7mm}{
		\begin{tabular}{@{\hskip 3mm}l@{\hskip 5mm}|@{\hskip 5mm}c@{\hskip 10mm}c@{\hskip 10mm}c@{\hskip 10mm}c@{\hskip 10mm}c@{\hskip 10mm}c@{\hskip 10mm}c}
		    \specialrule{.1em}{.05em}{.05em}
		    $\lambda$ &AP & AP$_{50}$ & AP$_{75}$ & AP$_{s}$ & AP$_{m}$ & AP$_{l}$ \\
		    \hline
            $1$ &42.1&\textbf{60.7}&\textbf{45.6}&\textbf{24.1}&44.5&58.3\\
            $10$ &\textbf{42.2}&60.6&45.5&23.9&\textbf{44.6}&\textbf{58.5}\\
            $100$ &41.7 & 60.3 & 44.9&23.4&44.1&57.5\\
            $300$ &41.0&59.4&44.3&22.2&43.5&57.2\\
	       \specialrule{.1em}{.05em}{.05em}
		\end{tabular}
		}
	\end{center}
	\caption{\textbf{Effect of the hyperparameter $\lambda$} in the selection loss  ($L_s=L_{iou}+\lambda L_{c}$) of DPP.}
	\vspace{-1em}
	\label{table:abl_lambda}
\end{table}

\noindent\textbf{Target number of heavy operators.} The target number $T$ of useage of heavy operator $g_0$ is leveraged in $L_c$ to control the complexity of DPP. $T$  is determined by two hyperparameters, the lower bound $T_{min}$ and the multiplier $\alpha$. Four DPP variants are implemented by varying these hyperparameters as shown in Figure \ref{fig:ddp_abl_dy}. The average number of times the operator $g_0$ is selected per image is $\{8,15,22,31\}$ in the COCO validation set, for models ranging from small to large FLOPS in Figure \ref{fig:ddp_abl_dy}. It can be seen that the precision of the model with $T_{min}=1$ and $\alpha=2$ is at the inflection point of the curve, beyond which the precision grows slowly for a large increase of the computational cost. When $\alpha=2$ and $T_{min}=1$, the model selects the $g_0$ operator $15$ times on average.  This is only twice the average instance number in COCO ($7$), confirming the effectiveness of loss function $L_c$. This result also suggests that using twice as many high complexity proposals as the number of object instances is a very effective choice in terms of the complexity-precision trade-off for the detection head.

In summary, the number of high complexity operators used on average can be very smaller than the overall number of proposals ($N=100$). Moreover, both hyperparameters can be used to modify the complexity of the detection head. The multiplier is more useful when the best complexity-precision trade-off is desired while the lower bound is more effective when the goal is to achieve the best precision irrespective of complexity.

\section{Conclusion}
In this paper, we propose to treat proposals of object detection unequally. A matching problem between proposals and operators is designed and optimized via a dynamic proposal processing (DPP) framework that contains a simple selector supervised with two loss functions, the IoU loss and the complexity loss. Experimental results show that the DPP framework achieves the state-of-the-art complexity-precision trade-off for the object detection on different types of backbones under a wide complexity range. We hope this paper can provide inspiration for different approaches of proposal processing by future research as well as research in deeper questions, such as the role of computational constraints in the development of effective vision systems.

\clearpage
%
%
\bibliographystyle{splncs04}
\bibliography{egbib}
\end{document}